# POMDPs under Probabilistic Semantics

**Krishnendu Chatterjee**
IST Austria
krishnendu.chatterjee@ist.ac.at

**Martin Chmelík**
IST Austria
martin.chmelik@ist.ac.at

## Abstract

We consider partially observable Markov decision processes (POMDPs) with limit-average payoff, where a reward value in the interval $[0,1]$ is associated to every transition, and the payoff of an infinite path is the long-run average of the rewards. We consider two types of path constraints: (i) quantitative constraint defines the set of paths where the payoff is at least a given threshold $\lambda_1 \in (0,1]$; and (ii) qualitative constraint which is a special case of quantitative constraint with $\lambda_1 = 1$. We consider the computation of the almost-sure winning set, where the controller needs to ensure that the path constraint is satisfied with probability 1. Our main results for qualitative path constraint are as follows: (i) the problem of deciding the existence of a finite-memory controller is EXPTIME-complete; and (ii) the problem of deciding the existence of an infinite-memory controller is undecidable. For quantitative path constraint we show that the problem of deciding the existence of a finite-memory controller is undecidable.

## 1 Introduction

**Partially observable Markov decision processes (POMDPs).** *Markov decision processes (MDPs)* are standard models for probabilistic systems that exhibit both probabilistic and nondeterministic behavior [10]. MDPs have been used to model and solve control problems for stochastic systems [7, 23]: nondeterminism represents the freedom of the controller to choose a control action, while the probabilistic component of the behavior describes the system response to control actions. In *perfect-observation (or perfect-information) MDPs (PIMDPs)* the controller can observe the current state of the system to choose the next control actions, whereas in *partially observable MDPs (POMDPs)* the state space is partitioned according to observations that the controller can observe, i.e., given the current state, the controller can only view the observation of the state (the partition the state belongs to), but not the precise state [20]. POMDPs provide the appropriate model to study a wide variety of applications such as in computational biology [5], speech processing [19], image processing [4], robot planning [13, 11], reinforcement learning [12], to name a few. POMDPs also subsume many other powerful computational models such as probabilistic finite automata (PFA) [24, 21] (since probabilistic finite automata (aka blind POMDPs) are a special case of POMDPs with a single observation).

**Limit-average payoff.** A *payoff* function maps every infinite path (infinite sequence of state action pairs) of a POMDP to a real value. The most well-studied payoff in the setting of POMDPs is the *limit-average* payoff where every state action pair is assigned a real-valued reward in the interval $[0,1]$ and the payoff of an infinite path is the long-run average of the rewards on the path [7, 23]. POMDPs with limit-average payoff provide the theoretical framework to study many important problems of practical relevance, including probabilistic planning and several stochastic optimization problems [11, 2, 16, 17, 27].

**Expectation vs probabilistic semantics.** Traditionally, MDPs with limit-average payoff have been studied with the *expectation* semantics, where the goal of the controller is to maximize the expected limit-average payoff. The expected payoff value can be $\frac{1}{2}$ when with probability $\frac{1}{2}$ the payoff is 1, and with remaining probability the payoff is 0. In many applications of system analysis (such as robot planning and control) the relevant question is the probability measure of the paths that satisfy certain criteria, e.g., whether the probability measure of the paths such that the limit-average payoff is 1 (or the payoff is at least

$\frac{1}{2}$) is at least a given threshold (e.g., see [1, 13]). We classify the path constraints for limit-average payoff as follows: (1) *quantitative constraint* that defines the set of paths with limit-average payoff at least $\lambda_1$, for a threshold $\lambda_1 \in (0, 1]$; and (2) *qualitative constraint* is the special case of quantitative constraint that defines the set of paths with limit-average payoff 1 (i.e., the special case with $\lambda_1 = 1$). We refer to the problem where the controller must satisfy a path constraint with a probability threshold $\lambda_2 \in (0, 1]$ as the *probabilistic* semantics. An important special case of probabilistic semantics is the *almost-sure* semantics, where the probability threshold is 1. The almost-sure semantics is of great importance because there are many applications where the requirement is to know whether the correct behavior arises with probability 1. For instance, when analyzing a randomized embedded scheduler, the relevant question is whether every thread progresses with probability 1. Even in settings where it suffices to satisfy certain specifications with probability $\lambda_2 < 1$, the correct choice of $\lambda_2$ is a challenging problem, due to the simplifications introduced during modeling. For example, in the analysis of randomized distributed algorithms it is quite common to require correctness with probability 1 (e.g., [22, 26]). Besides its importance in practical applications, almost-sure convergence, is a fundamental concept in probability theory, and provide stronger convergence guarantee than convergence in expectation [6].

**Previous results.** There are several deep undecidability results established for the special case of probabilistic finite automata (PFA) (that immediately imply undecidability for POMDPs). The basic undecidability results are for PFA over finite words: The emptiness problem for PFA under probabilistic semantics is undecidable over finite words [24, 21, 3]; and it was shown in [16] that even the following approximation version is undecidable: for any fixed $0 < \epsilon < \frac{1}{2}$, given a probabilistic automaton and the guarantee that either (a) there is a word accepted with probability at least $1-\epsilon$; or (ii) all words are accepted with probability at most $\epsilon$; decide whether it is case (i) or case (ii). The almost-sure problem for probabilistic automata over finite words reduces to the non-emptiness question of universal automata over finite words and is PSPACE-complete. However, another related decision question whether for every $\epsilon > 0$ there is a word that is accepted with probability at least $1 - \epsilon$ (called the value 1 problem) is undecidable for probabilistic automata over finite words [8]. Also observe that all undecidability results for probabilistic automata over finite words carry over to POMDPs where the controller is restricted to finite-memory strategies. The importance of finite-memory strategies in applications has been established in [9, 14, 18].

Table 1: Complexity: New results are in bold fonts

|  | Almost-sure semantics | | Prob. semantics |
|---|---|---|---|
|  | Fin. mem. | Inf. mem. | Fin./Inf. mem. |
| PFA | PSPACE-c | PSPACE-c | Undec. |
| POMDP Qual. Constr. | **EXPTIME-c** | Undec. | Undec. |
| POMDP Quan. Constr. | **Undec.** | **Undec.** | Undec. |

**Our contributions.** Since under the general probabilistic semantics, the decision problems are undecidable even for PFA, we consider POMDPs with limit-average payoff under the almost-sure semantics. We present a complete picture of decidability as well as optimal complexity.

*(Almost-sure winning for qualitative constraint).* We first consider limit-average payoff with qualitative constraint under almost-sure semantics. We show that *belief-based* strategies are not sufficient (where a belief-based strategy is based on the subset construction that remembers the possible set of current states): we show that there exist POMDPs with limit-average payoff with qualitative constraint where finite-memory almost-sure winning strategy exists but there exists no belief-based almost-sure winning strategy. Our counter-example shows that standard techniques based on subset construction (to construct an exponential size PIMDP) are not adequate to solve the problem. We then show one of our main result that given a POMDP with $|S|$ states and $|\mathcal{A}|$ actions, if there is a finite-memory almost-sure winning strategy to satisfy the limit-average payoff with qualitative constraint, then there is an almost-sure winning strategy that uses at most $2^{3 \cdot |S| + |\mathcal{A}|}$ memory. Our exponential memory upper bound is asymptotically optimal, as even for PFA over finite words, exponential memory is required for almost-sure winning (follows from the fact that the shortest witness word for non-emptiness of universal finite automata is at least exponential). We then show that the problem of deciding the existence of a finite-memory almost-sure winning strategy for limit-average payoff with qualitative constraint is EXPTIME-complete for POMDPs. In contrast to our result for finite-memory strategies, we show that deciding the existence of an infinite-memory almost-sure winning strategy for limit-average payoff with qualitative constraint is undecidable for POMDPs.

*(Almost-sure winning with quantitative constraint).* In contrast to our decidability result under finite-memory strategies for qualitative constraint, we show that the almost-sure winning problem for limit-average payoff with quantitative constraint is undecidable even for finite-memory strategies for POMDPs.

In summary we establish the precise decidability fron-tier for POMDPs with limit-average payoff under probabilistic semantics (see Table 1). For practical purposes, the most prominent question is the problem of finite-memory strategies, and for finite-memory strategies we establish decidability with EXPTIME-complete complexity for the important special case of qualitative constraint under almost-sure semantics.

## 2 Definitions

We present the definitions of POMDPs, strategies, objectives, and other basic notions required for our results. We follow standard notations from [23, 15].

**Notations.** Given a finite set $X$, we denote by $\mathcal{P}(X)$ the set of subsets of $X$, i.e., $\mathcal{P}(X)$ is the power set of $X$. A probability distribution $f$ on $X$ is a function $f: X \to [0,1]$ such that $\sum_{x \in X} f(x) = 1$, and we denote by $\mathcal{D}(X)$ the set of all probability distributions on $X$. For $f \in \mathcal{D}(X)$ we denote by $\mathrm{Supp}(f) = \{x \in X \mid f(x) > 0\}$ the support of $f$.

**Definition 1** (POMDP). *A Partially Observable Markov Decision Process (POMDP) is a tuple $G = (S, \mathcal{A}, \delta, \mathcal{O}, \gamma, s_0)$ where: (i) $S$ is a finite set of states; (ii) $\mathcal{A}$ is a finite alphabet of actions; (iii) $\delta: S \times \mathcal{A} \to \mathcal{D}(S)$ is a probabilistic transition function that given a state $s$ and an action $a \in \mathcal{A}$ gives the probability distribution over the successor states, i.e., $\delta(s,a)(s')$ denotes the transition probability from $s$ to $s'$ given action $a$; (iv) $\mathcal{O}$ is a finite set of observations; (v) $\gamma: S \to \mathcal{O}$ is an observation function that maps every state to an observation; and (vi) $s_0$ is the initial state.*

Given $s, s' \in S$ and $a \in \mathcal{A}$, we also write $\delta(s'|s,a)$ for $\delta(s,a)(s')$. A state $s$ is *absorbing* if for all actions $a$ we have $\delta(s,a)(s) = 1$ (i.e., $s$ is never left from $s$). For an observation $o$, we denote by $\gamma^{-1}(o) = \{s \in S \mid \gamma(s) = o\}$ the set of states with observation $o$. For a set $U \subseteq S$ of states and $O \subseteq \mathcal{O}$ of observations we denote $\gamma(U) = \{o \in \mathcal{O} \mid \exists s \in U. \gamma(s) = o\}$ and $\gamma^{-1}(O) = \bigcup_{o \in O} \gamma^{-1}(o)$.

**Plays and belief-updates.** A *play* (or a path) in a POMDP is an infinite sequence $(s_0, a_0, s_1, a_1, s_2, a_2, \ldots)$ of states and actions such that for all $i \geq 0$ we have $\delta(s_i, a_i)(s_{i+1}) > 0$. We write $\Omega$ for the set of all plays. For a finite prefix $w = (s_0, a_0, s_1, a_1, \ldots, s_n)$ we denote by $\gamma(w) = (\gamma(s_0), a_0, \gamma(s_1), a_1, \ldots, \gamma(s_n))$ the observation and action sequence associated with $w$. For a finite sequence $\rho = (o_0, a_0, o_1, a_1, \ldots, o_n)$ of observations and actions, the *belief* $\mathcal{B}(\rho)$ after the prefix $\rho$ is the set of states in which a finite prefix of a play can be after the sequence $\rho$ of observations and actions, i.e., $\mathcal{B}(\rho) = \{s_n = \mathsf{Last}(w) \mid w = (s_0, a_0, s_1, a_1, \ldots, s_n), w$ is a prefix of a play, and for all $0 \leq i \leq n. \gamma(s_i) = o_i\}$. The belief-updates associated with finite-prefixes are as follows: for prefixes $w$ and $w' = w \cdot a \cdot s$ the belief update is defined inductively $\mathcal{B}(\gamma(w')) = \left(\bigcup_{s_1 \in \mathcal{B}(\gamma(w))} \mathrm{Supp}(\delta(s_1, a))\right) \cap \gamma^{-1}(\gamma(s))$.

**Strategies.** A *strategy (or a policy)* is a recipe to extend prefixes of plays and is a function $\sigma: (S \cdot \mathcal{A})^* \cdot S \to \mathcal{D}(\mathcal{A})$ that given a finite history (i.e., a finite prefix of a play) selects a probability distribution over the actions. Since we consider POMDPs, strategies are *observation-based*, i.e., for all histories $w = (s_0, a_0, s_1, a_1, \ldots, a_{n-1}, s_n)$ and $w' = (s'_0, a_0, s'_1, a_1, \ldots, a_{n-1}, s'_n)$ such that for all $0 \leq i \leq n$ we have $\gamma(s_i) = \gamma(s'_i)$ (i.e., $\gamma(w) = \gamma(w')$), we must have $\sigma(w) = \sigma(w')$. In other words, if the observation sequence is the same, then the strategy cannot distinguish between the prefixes and must play the same. We now present an equivalent definition of observation-based strategies such that the memory of the strategy is explicitly specified, and will be required to present finite-memory strategies.

**Definition 2** (Strategies with memory and finite-memory strategies). *A strategy with memory is a tuple $\sigma = (\sigma_u, \sigma_n, M, m_0)$ where:(i) (Memory set). $M$ is a denumerable set (finite or infinite) of memory elements (or memory states). (ii) (Action selection function). The function $\sigma_n: M \to \mathcal{D}(\mathcal{A})$ is the action selection function that given the current memory state gives the probability distribution over actions. (iii) (Memory update function). The function $\sigma_u: M \times \mathcal{O} \times \mathcal{A} \to \mathcal{D}(M)$ is the memory update function that given the current memory state, the current observation and action, updates the memory state probabilistically. (iv) (Initial memory). The memory state $m_0 \in M$ is the initial memory state. A strategy is a finite-memory strategy if the set $M$ of memory elements is finite. A strategy is pure (or deterministic) if the memory update function and the action selection function are deterministic, i.e., $\sigma_u: M \times \mathcal{O} \times \mathcal{A} \to M$ and $\sigma_n: M \to \mathcal{A}$. A strategy is memoryless (or stationary) if it is independent of the history but depends only on the current observation, and can be represented as a function $\sigma: \mathcal{O} \to \mathcal{D}(\mathcal{A})$.*

**Objectives.** An *objective* in a POMDP $G$ is a measurable set $\varphi \subseteq \Omega$ of plays. We first define *limit-average payoff* (aka mean-payoff) function. Given a POMDP we consider a reward function $\mathsf{r}: S \times \mathcal{A} \to [0,1]$ that maps every state action pair to a real-valued reward in the interval $[0,1]$. The $\mathsf{LimAvg}$ payoff function maps every play to a real-valued reward that is the long-run average of the rewards of the play. Formally, given a play $\rho = (s_0, a_0, s_1, a_1, s_2, a_2, \ldots)$ we have $\mathsf{LimAvg}(\mathsf{r}, \rho) = \liminf_{n \to \infty} \frac{1}{n} \cdot \sum_{i=0}^{n} \mathsf{r}(s_i, a_i)$. When the reward function $\mathsf{r}$ is clear from the context, we drop it for simplicity. For a reward function r, we consider two types of limit-average payoff constraints: (i) *Qualitative constraint.* The *qualitative constraint* limit-average objective $\mathsf{LimAvg}_{=1}$ defines the set of paths such that the limit-average payoff is 1; i.e., $\mathsf{LimAvg}_{=1} = \{\rho \mid \mathsf{LimAvg}(\rho) = 1\}$. (ii) *Quantitative constraints.* Given a threshold $\lambda_1 \in (0,1)$, the *quantitative constraint* limit-average objective $\mathsf{LimAvg}_{>\lambda_1}$ defines the set of paths such that the limit-average payoff is strictly greater than $\lambda_1$; i.e., $\mathsf{LimAvg}_{>\lambda_1} = \{\rho \mid \mathsf{LimAvg}(\rho) > \lambda_1\}$.

**Probabilistic and almost-sure winning.** Given a POMDP, an objective $\varphi$, and a class $\mathcal{C}$ of strategies, we say that: (i) a strategy $\sigma \in \mathcal{C}$ is *almost-sure winning* if $\mathbb{P}^\sigma(\varphi) = 1$; (ii) a strategy $\sigma \in \mathcal{C}$ is *probabilistic winning*, for a threshold $\lambda_2 \in (0,1)$, if $\mathbb{P}^\sigma(\varphi) \geq \lambda_2$.

**Theorem 1** (Results for PFA (probabilistic automata over finite words) [21]). *The following assertions hold for the class $\mathcal{C}$ of all infinite-memory as well as finite-memory strategies: (1) the probabilistic winning problem is undecidable for PFA; and (2) the almost-sure winning problem is PSPACE-complete for PFA.*

Since PFA are a special case of POMDPs, the undecidability of the probabilistic winning problem for PFA implies the undecidability of the probabilistic winning problem for POMDPs with both qualitative and quantitative constraint limit-average objectives. The almost-sure winning problem is PSPACE-complete for PFAs, and we study the complexity of the almost-sure winning problem for POMDPs with both qualitative and quantitative constraint limit-average objectives, under infinite-memory and finite-memory strategies.

**Basic properties of Markov Chains.** Since our proofs will use results of Markov chains, we start with some basic results related to Markov chains.

*Markov chains and recurrent classes.* A Markov chain $\overline{G} = (\overline{S}, \overline{\delta})$ consists of a finite set $\overline{S}$ of states and a probabilistic transition function $\overline{\delta} : \overline{S} \to \mathcal{D}(\overline{S})$. Given the Markov chain, we consider the directed graph $(\overline{S}, \overline{E})$ where $\overline{E} = \{(\overline{s}, \overline{s}') \mid \delta(\overline{s}' \mid \overline{s}) > 0\}$. A *recurrent class* $\overline{C} \subseteq \overline{S}$ of the Markov chain is a bottom strongly connected component (scc) in the graph $(\overline{S}, \overline{E})$ (a bottom scc is an scc with no edges out of the scc). We denote by $\mathsf{Rec}(\overline{G})$ the set of recurrent classes of the Markov chain, i.e., $\mathsf{Rec}(\overline{G}) = \{\overline{C} \mid \overline{C}$ is a recurrent class$\}$. Given a state $\overline{s}$ and a set $\overline{U}$ of states, we say that $\overline{U}$ is reachable from $\overline{s}$ if there is a path from $\overline{s}$ to some state in $\overline{U}$ in the graph $(\overline{S}, \overline{E})$. Given a state $\overline{s}$ of the Markov chain we denote by $\mathsf{Rec}(\overline{G})(\overline{s}) \subseteq \mathsf{Rec}(\overline{G})$ the subset of the recurrent classes reachable from $\overline{s}$ in $\overline{G}$. A state is *recurrent* if it belongs to a recurrent class. The following standard properties of reachability and the recurrent classes will be used in our proofs:

- *Property 1.* (a) For a set $\overline{T} \subseteq \overline{S}$, if for all states $\overline{s} \in \overline{S}$ there is a path to $\overline{T}$ (i.e., for all states there is a positive probability to reach $\overline{T}$), then from all states the set $\overline{T}$ is reached with probability 1. (b) For all states $\overline{s}$, if the Markov chain starts at $\overline{s}$, then the set $\overline{\mathsf{C}} = \bigcup_{\overline{C} \in \mathsf{Rec}(\overline{G})(\overline{s})} \overline{C}$ is reached with probability 1, i.e., the set of recurrent classes is reached with probability 1.
- *Property 2.* For a recurrent class $\overline{C}$, for all states $\overline{s} \in \overline{C}$, if the Markov chain starts at $\overline{s}$, then for all states $\overline{t} \in \overline{C}$ the state $\overline{t}$ is visited infinitely often with probability 1, and is visited with positive average frequency (i.e., positive limit-average frequency) with probability 1.

Lemma 1 is a consequence of the above properties.

**Lemma 1.** *Let $\overline{G} = (\overline{S}, \overline{\delta})$ be a Markov chain with a reward function $\mathsf{r} : \overline{S} \to [0,1]$, and $\overline{s} \in \overline{S}$ a state of the Markov chain. The state $\overline{s}$ is almost-sure winning for the objective $\mathsf{LimAvg}_{=1}$ iff for all recurrent classes $\overline{C} \in \mathsf{Rec}(\overline{G})(\overline{s})$ and for all states $\overline{s}_1 \in \overline{C}$ we have $\mathsf{r}(\overline{s}_1) = 1$.*

*Markov chains under finite memory strategies.* We now define Markov chains obtained by fixing a finite-memory strategy in a POMDP $G$. A finite-memory strategy $\sigma = (\sigma_u, \sigma_n, M, m_0)$ induces a Markov chain $(S \times M, \delta_\sigma)$, denoted $G\!\!\restriction_\sigma$, with the probabilistic transition function $\delta_\sigma : S \times M \to \mathcal{D}(S \times M)$: given $s, s' \in S$ and $m, m' \in M$, the transition $\delta_\sigma\big((s', m') \mid (s, m)\big)$ is the probability to go from state $(s, m)$ to state $(s', m')$ in one step under the strategy $\sigma$. The probability of transition can be decomposed as follows: (i) First an action $a \in \mathcal{A}$ is sampled according to the distribution $\sigma_n(m)$; (ii) then the next state $s'$ is sampled according to the distribution $\delta(s, a)$; and (iii) finally the new memory $m'$ is sampled according to the distribution $\sigma_u(m, \gamma(s'), a)$ (i.e., the new memory is sampled according $\sigma_u$ given the old memory, new observation and the action). More formally, we have: $\delta_\sigma\big((s', m') \mid (s, m)\big) = \sum_{a \in \mathcal{A}} \sigma_n(m)(a) \cdot \delta(s, a)(s') \cdot \sigma_u(m, \gamma(s'), a)(m')$.

## 3 Finite-memory strategies with Qualitative Constraint

In this section we show the following three results for finite-memory strategies: (i) in POMDPs with $\mathsf{LimAvg}_{=1}$ objectives belief-based strategies are not sufficient for almost-sure winning; (ii) an exponential upper bound on the memory required by an almost-sure winning strategy for $\mathsf{LimAvg}_{=1}$ objectives; and (iii) the decision problem is EXPTIME-complete.

**Belief is not sufficient.** We now show with an example that there exist POMDPs with $\mathsf{LimAvg}_{=1}$ objectives, where finite-memory randomized almost-sure winning strategies exist, but there exists no belief-based randomized almost-sure winning strategy (a belief-based strategy only uses memory that relies on the subset construction where the subset denotes the possible current states called belief). We will present the counter-example even for POMDPs with restricted reward function r assigning only Boolean rewards 0 and 1 to the states (does not depend on the action).

**Example 1.** *We consider a POMDP with state space $\{s_0, X, X', Y, Y', Z, Z'\}$ and action set $\{a, b\}$, and let $U = \{X, X', Y, Y', Z, Z'\}$. From the initial state $s_0$ all the other states are reached with uniform probability in one-step, i.e., for all $s' \in U = \{X, X', Y, Y', Z, Z'\}$ we have $\delta(s_0, a)(s') = \delta(s_0, b)(s') = \frac{1}{6}$. The transitions from the other states are shown in Figure 1. All states in $U$ have the same observation. The reward function r assigns the reward 1 to states $X, X', Z, Z'$ and 0 to states $Y$ and $Y'$. The belief initially after one-step is the set $U = \{X, X', Y, Y', Z, Z'\}$ since from $s_0$ all of them are reached with positive probability. The belief is always the set $U$ since every state has an input edge for every action, i.e., if the current belief is $U$ (the set of states that the POMDP is currently in with positive probability is $U$), then irrespective of whether $a$ or $b$ is chosen all states of $U$ are reached with positive probability and thus the belief is again $U$. There are three belief-based strategies: (i) $\sigma_1$ that plays always a; (ii) $\sigma_2$ that plays always b; or (iii) $\sigma_3$ that plays both a and b with positive probability. The Markov chains $G\!\upharpoonright_{\sigma_1}$ and $G\!\upharpoonright_{\sigma_2}$ are also shown in Figure 1, and the graph of $G\!\upharpoonright_{\sigma_3}$ is the same as the POMDP $G$ (with edge labels removed). For all the three strategies, the Markov chains contain the whole set $U$ as the reachable recurrent class, and it follows by Lemma 1 that none of the belief-based strategies $\sigma_1, \sigma_2$ or $\sigma_3$ are almost-sure winning for the $\mathsf{LimAvg}_{=1}$ objective. The strategy $\sigma_4$ that plays action a and b alternately gives rise to a Markov chain where the recurrent classes do not intersect with $Y$ or $Y'$, and is a finite-memory almost-sure winning strategy for the $\mathsf{LimAvg}_{=1}$ objective.* □

### 3.1 Strategy complexity

For the rest of the subsection we fix a finite-memory almost-sure winning strategy $\sigma = (\sigma_u, \sigma_n, M, m_0)$ on the POMDP $G = (S, \mathcal{A}, \delta, \mathcal{O}, \gamma, s_0)$ with a reward function r for the objective $\mathsf{LimAvg}_{=1}$. Our goal is to construct an almost-sure winning strategy for the $\mathsf{LimAvg}_{=1}$ objective with memory size at most $\mathsf{Mem}^* = 2^{3 \cdot |S|} \cdot 2^{|\mathcal{A}|}$. We start with a few definitions associated with strategy $\sigma$. For $m \in M$:

- The function $\mathsf{RecFun}_\sigma(m) : S \to \{0,1\}$ is such that $\mathsf{RecFun}_\sigma(m)(s)$ is 1 iff the state $(s, m)$ is recurrent in the Markov chain $G\!\upharpoonright_\sigma$ and 0 otherwise.
- The function $\mathsf{AWFun}_\sigma(m) : S \to \{0,1\}$ is such

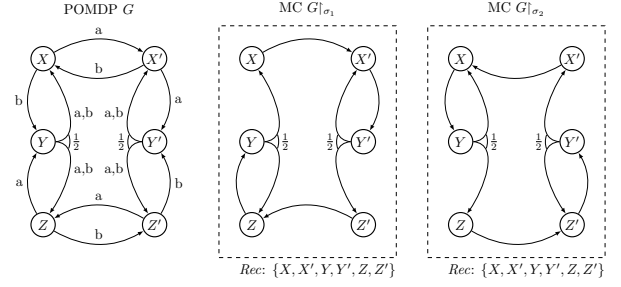

Figure 1: Belief is not sufficient

that $\mathsf{AWFun}_\sigma(m)(s)$ is 1 iff the state $(s, m)$ is almost-sure winning for the $\mathsf{LimAvg}_{=1}$ objective in the Markov chain $G\!\upharpoonright_\sigma$ and 0 otherwise.
- We also consider $\mathsf{Act}_\sigma(m) = \mathsf{Supp}(\sigma_n(m))$ that for every memory element gives the support of the probability distribution over actions played at $m$.

**Remark 1.** *Let $(s', m')$ be reachable from $(s, m)$ in $G\!\upharpoonright_\sigma$. If the state $(s, m)$ is almost-sure winning for the $\mathsf{LimAvg}_{=1}$ objective, then the state $(s', m')$ is also almost-sure winning for the $\mathsf{LimAvg}_{=1}$ objective.*

**Collapsed graph of $\sigma$.** Given the strategy $\sigma$ we define the notion of a *collapsed graph* $\mathsf{CoGr}(\sigma) = (V, E)$. The states of the graph are elements from the set $V = \{(Y, \mathsf{AWFun}_\sigma(m), \mathsf{RecFun}_\sigma(m), \mathsf{Act}_\sigma(m)) \mid Y \subseteq S \text{ and } m \in M\}$ and the initial state is $(\{s_0\}, \mathsf{AWFun}_\sigma(m_0), \mathsf{RecFun}_\sigma(m_0), \mathsf{Act}_\sigma(m_0))$. The edges in $E$ are labeled by actions in $\mathcal{A}$. Intuitively, the action labeled edges of the graph depict the updates of the belief and the functions upon a particular action. Formally, there is an edge $(Y, \mathsf{AWFun}_\sigma(m), \mathsf{RecFun}_\sigma(m), \mathsf{Act}_\sigma(m)) \xrightarrow{a} (Y', \mathsf{AWFun}_\sigma(m'), \mathsf{RecFun}_\sigma(m'), \mathsf{Act}_\sigma(m'))$ in the collapsed graph $\mathsf{CoGr}(\sigma)$ iff there exists an observation $o \in \mathcal{O}$ such that (i) the action $a \in \mathsf{Act}_\sigma(m)$; (ii) the set $Y'$ is non-empty and it is the belief update from $Y$, under action $a$ and the observation $o$, i.e., $Y' = \bigcup_{s \in Y} \mathsf{Supp}(\delta(s, a)) \cap \gamma^{-1}(o)$; and (iii) $m' \in \mathsf{Supp}(\sigma_u(m, o, a))$. Note that the number of states in the graph is bounded by $|V| \leq 2^{3 \cdot |S|} \cdot 2^{|\mathcal{A}|} = \mathsf{Mem}^*$.

We now define the collapsed strategy for $\sigma$. Intuitively we collapse memory elements of $\sigma$ whenever they agree on all the $\mathsf{RecFun}, \mathsf{AWFun},$ and $\mathsf{Act}$ functions. The collapsed strategy plays uniformly all the actions from the set given by $\mathsf{Act}$ in the collapsed state.

**Collapsed strategy.** We now construct the *collapsed strategy* $\sigma' = (\sigma'_u, \sigma'_n, M', m'_0)$ of $\sigma$ based on the collapsed graph $\mathsf{CoGr}(\sigma) = (V, E)$. We will refer to this construction by $\sigma' = \mathsf{CoSt}(\sigma)$.

- The memory set $M'$ are the vertices of the collapsed graph $\mathsf{CoGr}(\sigma) = (V, E)$, i.e., $M' = V = $

- $\{(Y, \mathsf{AWFun}_\sigma(m), \mathsf{RecFun}_\sigma(m), \mathsf{Act}_\sigma(m)) \mid Y \subseteq S \text{ and } m \in M\}$. Note that $|M'| \leq \mathsf{Mem}^*$.
- The initial memory is $m'_0 = (\{s_0\}, \mathsf{AWFun}_\sigma(m_0), \mathsf{RecFun}_\sigma(m_0), \mathsf{Act}_\sigma(m_0))$.
- The next action function given a memory $(Y, W, R, A) \in M'$ is the uniform distribution over the set of actions $\{a \mid \exists (Y', W', R', A') \in M' \text{ and } (Y, W, R, A) \overset{a}{\to} (Y', W', R', A') \in E\}$, where $E$ are the edges of the collapsed graph.
- The memory update function $\sigma'_u((Y, W, R, A), o, a)$ given a memory element $(Y, W, R, A) \in M'$, $a \in \mathcal{A}$, and $o \in \mathcal{O}$ is the uniform distribution over the set of states $\{(Y', W', R', A') \mid (Y, W, R, A) \overset{a}{\to} (Y', W', R', A') \in E \text{ and } Y' \subseteq \gamma^{-1}(o)\}$.

**Random variable notation.** For all $n \geq 0$ we write $X_n, Y_n, W_n, R_n, A_n, L_n$ for the random variables that correspond to the projection of the $n^{th}$ state of the Markov chain $G\!\upharpoonright_{\sigma'}$ on the $S$ component, the belief $\mathcal{P}(S)$ component, the $\mathsf{AWFun}_\sigma$ component, the $\mathsf{RecFun}_\sigma$ component, the $\mathsf{Act}_\sigma$ component, and the $n^{th}$ action, respectively.

**Run of the Markov chain $G\!\upharpoonright_{\sigma'}$.** A *run* of the Markov chain $G\!\upharpoonright_{\sigma'}$ is an infinite sequence

$$(X_0, Y_0, W_0, R_0, A_0) \overset{L_0}{\to} (X_1, Y_1, W_1, R_1, A_1) \overset{L_1}{\to} \cdots$$

such that each finite prefix of the run is generated with positive probability on the Markov chain, i.e., for all $i \geq 0$, we have (i) $L_i \in \mathsf{Supp}(\sigma'_n(Y_i, W_i, R_i, A_i))$; (ii) $X_{i+1} \in \mathsf{Supp}(\delta(X_i, L_i))$; and (iii) $(Y_{i+1}, W_{i+1}, R_{i+1}, A_{i+1}) \in \mathsf{Supp}(\sigma'_u((Y_i, W_i, R_i, A_i), \gamma(X_{i+1}), L_i))$. In the following lemma we establish important properties of the Markov chain $G\!\upharpoonright_{\sigma'}$ that are essential for our proof.

**Lemma 2.** *Let* $(X_0, Y_0, W_0, R_0, A_0) \overset{L_0}{\to} (X_1, Y_1, W_1, R_1, A_1) \overset{L_1}{\to} \cdots$ *be a run of the Markov chain* $G\!\upharpoonright_{\sigma'}$, *then the following assertions hold for all* $i \geq 0$: *1.* $X_{i+1} \in \mathsf{Supp}(\delta(X_i, L_i)) \cap Y_{i+1}$; *2.* $(Y_i, W_i, R_i, A_i) \overset{L_i}{\to} (Y_{i+1}, W_{i+1}, R_{i+1}, A_{i+1})$ *is an edge in the collapsed graph* $\mathsf{CoGr}(\sigma)$; *3. if* $W_i(X_i) = 1$, *then* $W_{i+1}(X_{i+1}) = 1$; *4. if* $R_i(X_i) = 1$, *then* $R_{i+1}(X_{i+1}) = 1$; *and 5. if* $W_i(X_i) = 1$ *and* $R_i(X_i) = 1$, *then* $\mathsf{r}(X_i, L_i) = 1$.

*Proof.* We present the proof of the fifth point, and the other points are straight-forward. For the fifth point consider that $W_i(X_i) = 1$ and $R_i(X_i) = 1$. Then there exists a memory $m \in M$ such that (i) $\mathsf{AWFun}_\sigma(m) = W_i$, and (ii) $\mathsf{RecFun}_\sigma(m) = R_i$. Moreover, the state $(X_i, m)$ is a recurrent (since $R_i(X_i) = 1$) and almost-sure winning state (since $W_i(X_i) = 1$) in the Markov chain $G\!\upharpoonright_\sigma$. As $L_i \in \mathsf{Act}_\sigma(m)$ it follows that $L_i \in \mathsf{Supp}(\sigma_n(m))$, i.e., the action $L_i$ is played with positive probability in state $X_i$ given memory $m$, and $(X_i, m)$ is in an almost-sure winning recurrent class. By Lemma 1 it follows that the reward $\mathsf{r}(X_i, L_i)$ must be 1. The desired result follows. □

We now introduce the final notion of a collapsed-recurrent state that is required to complete the proof. A state $(X, Y, W, R, A)$ of the Markov chain $G\!\upharpoonright_{\sigma'}$ is collapsed-recurrent, if for all memory elements $m \in M$ that were merged to the memory element $(Y, W, R, A)$, the state $(X, m)$ of the Markov chain $G\!\upharpoonright_\sigma$ is recurrent. It will turn out that every recurrent state of the Markov chain $G\!\upharpoonright_\sigma$ is also collapsed-recurrent.

**Definition 3.** *A state* $(X, Y, W, R, A)$ *of the Markov chain* $G\!\upharpoonright_{\sigma'}$ *is called collapsed-recurrent iff* $R(X) = 1$.

Note that due to point 4 of Lemma 2 all the states reachable from a collapsed-recurrent state are also collapsed-recurrent. In the following lemma we show that the set of collapsed-recurrent states is reached with probability 1; and the key fact we show is that from every state in $G\!\upharpoonright_{\sigma'}$ a collapsed-recurrent state is reached with positive probability, and then use Property 1 (a) of Markov chains to establish the lemma.

**Lemma 3.** *With probability 1 a run of the Markov chain* $G\!\upharpoonright_{\sigma'}$ *reaches a collapsed-recurrent state.*

**Lemma 4.** *The collapsed strategy* $\sigma'$ *is a finite-memory almost-sure winning strategy for the* $\mathsf{LimAvg}_{=1}$ *objective on the POMDP* $G$ *with the reward function* $\mathsf{r}$.

*Proof.* The initial state of the Markov chain $G\!\upharpoonright_{\sigma'}$ is $(\{s_0\}, \mathsf{AWFun}_\sigma(m_0), \mathsf{RecFun}_\sigma(m_0), \mathsf{Act}_\sigma(m_0))$ and as the strategy $\sigma$ is an almost-sure winning strategy we have that $\mathsf{AWFun}_\sigma(m_0)(s_0) = 1$. It follows from the third point of Lemma 2 that every reachable state $(X, Y, W, R, A)$ in the Markov chain $G\!\upharpoonright_{\sigma'}$ satisfies that $W(X) = 1$. From the initial state a collapsed-recurrent state is reached with probability 1. It follows that all the recurrent states in the Markov chain $G\!\upharpoonright_{\sigma'}$ are also collapsed-recurrent states. As in all reachable states $(X, Y, W, R, A)$ we have $W(X) = 1$, by the fifth point of Lemma 2 it follows that every action $L$ played in a collapsed-recurrent state $(X, Y, W, R, A)$ satisfies that the reward $\mathsf{r}(X, L) = 1$. As this true for every reachable recurrent class, the fact that the collapsed strategy is an almost-sure winning strategy for $\mathsf{LimAvg}_{=1}$ objective follows from Lemma 1. □

**Theorem 2** (Strategy complexity). *The following assertions hold: (1) If there exists a finite-memory almost-sure winning strategy in the POMDP* $G = (S, \mathcal{A}, \delta, \mathcal{O}, \gamma, s_0)$ *with reward function* $\mathsf{r}$ *for the* $\mathsf{LimAvg}_{=1}$ *objective, then there exists a finite-memory* almost-sure winning strategy with memory size at most $2^{3\cdot|S|+|\mathcal{A}|}$. (2) Finite-memory almost-sure winning strategies for $\mathsf{LimAvg}_{=1}$ objectives in POMDPs in general require exponential memory and belief-based strategies are not sufficient.

*Proof.* The first point follows from Lemma 4 and the fact that the size of the memory set of the collapsed strategy $\sigma'$ of any finite-memory strategy $\sigma$ (which is the size of the vertex set of the collapsed graph of $\sigma$) is bounded by $2^{3\cdot|S|+|\mathcal{A}|}$. □

### 3.2 Computational complexity

A naive double-exponential time algorithm would be to enumerate all finite-memory strategies with memory bounded by $2^{3\cdot|S|+|\mathcal{A}|}$ (by Theorem 2). Our improved exponential-time algorithm consists of two steps: (i) first it constructs a special type of a *belief-observation* POMDP $\overline{G}$ from a POMDP $G$ (and $\overline{G}$ is exponential in $G$); and we show that there exists a finite-memory almost-sure winning strategy for the objective $\mathsf{LimAvg}_{=1}$ in $G$ iff there exists a randomized memoryless almost-sure winning strategy in $\overline{G}$ for the objective $\mathsf{LimAvg}_{=1}$; and (ii) then we show how to determine whether there exists a randomized memoryless almost-sure winning strategy in $\overline{G}$ for the $\mathsf{LimAvg}_{=1}$ objective in polynomial time with respect to the size of $\overline{G}$. For a belief-observation POMDP the current belief is always the set of states with current observation.

**Definition 4.** *A POMDP $\overline{G} = (\overline{S}, \overline{\mathcal{A}}, \overline{\delta}, \overline{\mathcal{O}}, \overline{\gamma}, \overline{s}_0)$ is a* belief-observation *POMDP iff for every finite prefix $\overline{w} = (\overline{s}_0, \overline{a}_0, \overline{s}_1, \overline{a}_1, \ldots, \overline{a}_{n-1}, \overline{s}_n)$ the belief associated with the observation sequence $\overline{\rho} = \overline{\gamma}(\overline{w})$ is the set of states with the last observation $\overline{\gamma}(\overline{s}_n)$ of the observation sequence $\overline{\rho}$, i.e., $\mathcal{B}(\overline{\rho}) = \overline{\gamma}^{-1}(\overline{\gamma}(\overline{s}_n))$.*

**Construction of the belief-observation POMDP.** Intuitively, the construction of $\overline{G}$ from $G$ will proceed as follows: if there exists an almost-sure winning finite-memory strategy, then there exists an almost-sure winning collapsed strategy with memory bounded by $2^{3\cdot|S|+|\mathcal{A}|}$. This allows us to consider the memory elements $M = 2^S \times \{0,1\}^{|S|} \times \{0,1\}^{|S|} \times 2^{\mathcal{A}}$; and intuitively construct the product of the memory $M$ with the POMDP $G$. The POMDP $\overline{G}$ is constructed such that it allows all possible ways the collapsed strategy of a finite-memory almost-sure winning strategy could play. The reward function $\overline{r}$ in $\overline{G}$ is obtained from the reward function $r$ in $G$. In the POMDP $\overline{G}$ the belief is already included in the state space itself of the POMDP, and the belief represents exactly the set of states in which the POMDP can be with positive probability. Therefore, the POMDP $\overline{G}$ is a belief-observation POMDP. Since possible memory states of collapsed strategies are part of state space, we only need to consider memoryless strategies in $\overline{G}$.

**Lemma 5.** *The POMDP $\overline{G}$ is a belief-observation POMDP, such that there exists a finite-memory almost-sure winning strategy for the $\mathsf{LimAvg}_{=1}$ objective with the reward function $r$ in the POMDP $G$ iff there exists a memoryless almost-sure winning strategy for the $\mathsf{LimAvg}_{=1}$ objective with the reward function $\overline{r}$ in the POMDP $\overline{G}$.*

**Almost-sure winning observations.** Given a POMDP $\overline{G} = (\overline{S}, \overline{\mathcal{A}}, \overline{\delta}, \overline{\mathcal{O}}, \overline{\gamma})$ and an objective $\psi$, let $\mathsf{Almost}_{\mathcal{M}}(\psi)$ denote the set of observations $\overline{o} \in \overline{\mathcal{O}}$, such that there exists a memoryless almost-sure winning strategy to ensure $\psi$ from every state $\overline{s} \in \overline{\gamma}^{-1}(\overline{o})$.

**Almost-sure winning for $\mathsf{LimAvg}_{=1}$ objectives.** Our goal is to compute the set $\mathsf{Almost}_{\mathcal{M}}(\mathsf{LimAvg}_{=1})$ given the belief-observation POMDP $\overline{G}$ (of our construction of $G$ with product with $M$). Let $F \subseteq \overline{S}$ be the set of states of $\overline{G}$ where some actions can be played consistently with the collapsed strategy of any finite-memory almost-sure winning strategy. Let $\widetilde{G}$ denote the POMDP $\overline{G}$ restricted to $F$. We define a subset of states of the belief-observation POMDP $\widetilde{G}$ that intuitively correspond to winning collapsed-recurrent states (wcs), i.e., $\widetilde{S}_{wcs} = \{(s, (Y, W, R, A)) \mid W(s) = 1, R(s) = 1\}$. Then, we compute the set of observations $\widetilde{\mathsf{AW}}$ that can ensure to reach $\widetilde{S}_{wcs}$ almost-surely in the the POMDP $\widetilde{G}$. We show that the set of observations $\widetilde{\mathsf{AW}}$ is equal to the set of observations $\mathsf{Almost}_{\mathcal{M}}(\mathsf{LimAvg}_{=1})$ in the POMDP $\overline{G}$. Thus the computation reduces to computation of almost-sure states for reachability objectives. Finally we show that almost-sure reachability set can be computed in quadratic time for belief-observation POMDPs. The quadratic time algorithm is obtained as follows: we show almost-sure winning observations to ensure to reach a target set $\overline{T}$ with probability 1 is the greatest fixpoint of a set $\overline{Y}$ of observations such that playing all actions uniformly that ensures $\overline{Y}$ is not left, ensures to reach $\overline{T}$ almost-surely. This characterization gives a nested iterative algorithm that is quadratic time.

**Lemma 6.** $\widetilde{\mathsf{AW}} = \mathsf{Almost}_{\mathcal{M}}(\mathsf{LimAvg}_{=1})$; *and $\widetilde{\mathsf{AW}}$ can be computed in quadratic time in the size of $\overline{G}$.*

**The EXPTIME-completeness.** In this section we first showed that given a POMDP $G$ with a $\mathsf{LimAvg}_{=1}$ objective we can construct an exponential size belief-observation POMDP $\overline{G}$ and the computation of the almost-sure winning set for $\mathsf{LimAvg}_{=1}$ objectives is reduced to the computation of the almost-sure winning set for reachability objectives, which we solve in quadratic time in $\overline{G}$. This gives us an exponential-time algorithm to decide (and construct if one ex-ists) the existence of finite-memory almost-sure winning strategies in POMDPs with $\mathsf{LimAvg}_{=1}$ objectives. The EXPTIME-hardness for almost-sure winning can be obtained easily from the result of Reif for two-player partial-observation games with safety objectives [25].

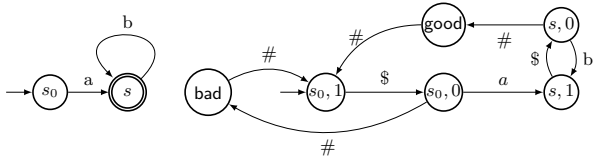

Figure 2: PFA P to a POMDP G

**Theorem 3.** *The following assertions hold: (1) Given a POMDP G with $|S|$ states, $|\mathcal{A}|$ actions, and a $\mathsf{LimAvg}_{=1}$ objective, the existence (and construction if one exists) of a finite-memory almost-sure winning strategy can be achieved in $2^{O(|S|+|\mathcal{A}|)}$ time. (2) The decision problem of given a POMDP and a $\mathsf{LimAvg}_{=1}$ objective whether there exists a finite-memory almost-sure winning strategy is EXPTIME-complete.*

**Remark 2.** *We considered observation function that assigns an observation to every state. In general the observation function $\gamma : S \to 2^{\mathcal{O}} \setminus \emptyset$ may assign multiple observations to a single state. In that case we consider the set of observations as $\mathcal{O}' = 2^{\mathcal{O}} \setminus \emptyset$ and consider the mapping that assigns to every state an observation from $\mathcal{O}'$ and then apply our results.*

## 4 Finite-memory strategies with Quantitative Constraint

We will show that the problem of deciding whether there exists a finite-memory (as well as an infinite-memory) almost-sure winning strategy for the objective $\mathsf{LimAvg}_{>\frac{1}{2}}$ is undecidable. We present a reduction from the standard undecidable problem for probabilistic finite automata (PFA). A PFA $\mathsf{P} = (S, \mathcal{A}, \delta, F, s_0)$ is a special case of a POMDP $G = (S, \mathcal{A}, \delta, \mathcal{O}, \gamma, s_0)$ with a single observation $\mathcal{O} = \{o\}$ such that for all states $s \in S$ we have $\gamma(s) = o$. Moreover, the PFA proceeds for only finitely many steps, and has a set $F$ of desired final states. The *strict emptiness problem* asks for the existence of a strategy $w$ (a finite word over the alphabet $\mathcal{A}$) such that the measure of the runs ending in the desired final states $F$ is strictly greater than $\frac{1}{2}$; and the strict emptiness problem for PFA is undecidable [21].

**Reduction.** Given a PFA $\mathsf{P} = (S, \mathcal{A}, \delta, F, s_0)$ we construct a POMDP $G = (S', \mathcal{A}', \delta', \mathcal{O}, \gamma, s_0')$ with a Boolean reward function $\mathsf{r}$ such that there exists a word $w \in \mathcal{A}^*$ accepted with probability strictly greater than $\frac{1}{2}$ in P iff there exists a finite-memory almost-sure winning strategy in $G$ for the objective $\mathsf{LimAvg}_{>\frac{1}{2}}$. Intuitively, the construction of the POMDP $G$ is as follows: for every state $s \in S$ of P we construct a pair of states $(s, 1)$ and $(s, 0)$ in $S'$ with the property that $(s, 0)$ can only be reached with a new action \$ (not in $\mathcal{A}$) played in state $(s, 1)$. The transition function $\delta'$ from the state $(s, 0)$ mimics the transition function $\delta$, i.e., $\delta'((s, 0), a)((s', 1)) = \delta(s, a)(s')$. The reward r of $(s, 1)$ (resp. $(s, 0)$) is 1 (resp. 0), ensuring the average of the pair to be $\frac{1}{2}$. We add a new available action # that when played in a final state reaches a state $\mathsf{good} \in S'$ with reward 1, and when played in a non-final state reaches a state $\mathsf{bad} \in S'$ with reward 0, and for states $\mathsf{good}$ and $\mathsf{bad}$ given action # the next state is the initial state. An illustration of the construction on an example is depicted on Figure 2. Whenever an action is played in a state where it is not available, the POMDP reaches a loosing absorbing state, i.e., an absorbing state with reward 0, and for brevity we omit transitions to the loosing absorbing state. We present key proof ideas to establish the correctness:

*(Strict emptiness implies almost-sure $\mathsf{LimAvg}_{>\frac{1}{2}}$).* Let $w \in \mathcal{A}^*$ be a word accepted in P with probability $\mu > \frac{1}{2}$ and let the length of the word be $|w| = n$. We construct a pure finite-memory almost-sure winning strategy for the objective $\mathsf{LimAvg}_{>\frac{1}{2}}$ in the POMDP $G$ as follows: We denote by $w[i]$ the $i^{th}$ action in the word $w$. The finite-memory strategy we construct is specified as an ultimately periodic word $(\$w[1]\$w[2]\ldots\$w[n]\#\#)^\omega$. Observe that by the construction of the POMDP $G$, the sequence of rewards (that appear on the transitions) is $(10)^n$ followed by (i) 1 with probability $\mu$ (when $F$ is reached), and (ii) 0 otherwise; and the whole sequence is repeated ad infinitum. Then using the Strong Law of Large Numbers (SLLN) [6, Theorem 7.1, page 56] we show that with probability 1 the objective $\mathsf{LimAvg}_{>\frac{1}{2}}$ is satisfied.

*(Almost-sure $\mathsf{LimAvg}_{>\frac{1}{2}}$ implies strict emptiness).* Conversely, if there is a pure finite-memory strategy $\sigma$ to ensure the objective $\mathsf{LimAvg}_{>\frac{1}{2}}$ in the POMDP, then the strategy $\sigma$ can be viewed as an ultimately periodic infinite word of the form $u \cdot v^\omega$, where $u, v$ are finite words from $\mathcal{A}'$. Note that $v$ must contain the subsequence ##, as otherwise the $\mathsf{LimAvg}$ payoff would be only $\frac{1}{2}$. Similarly, before every letter $a \in \mathcal{A}$ in the words $u, v$, the strategy must necessarily play the \$ action, as otherwise the loosing absorbing state is reached. Again using SLLN we show that from the word $v$ we can extract a word $w$ that is accepted in the PFA with probability strictly greater than $\frac{1}{2}$. Finally, we show that if there is randomized (possibly infinite-memory strategy) to ensure the objective $\mathsf{LimAvg}_{>\frac{1}{2}}$ in the POMDP, then there is a pure finite-memory strategy as well (the technical proof uses Fa-tou's lemma [6]).

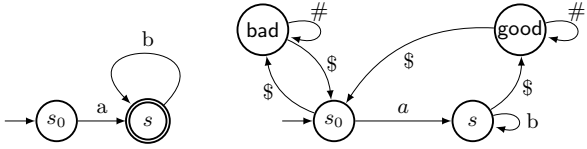

Figure 3: PFA P to a POMDP $G$

**Theorem 4.** *The problem whether there exists a finite (or infinite-memory) almost-sure winning strategy in a POMDP for the objective* $\mathsf{LimAvg}_{>\frac{1}{2}}$ *is undecidable.*

## 5 Infinite-memory strategies with Qualitative Constraint

In this section we show that the problem of deciding the existence of infinite-memory almost-sure winning strategies in POMDPs with $\mathsf{LimAvg}_{=1}$ objectives is undecidable. We prove this fact by a reduction from the *value 1 problem* in PFA, which is undecidable [8]. The value 1 problem given a PFA P asks whether for every $\epsilon > 0$ there exists a finite word $w$ such that the word is accepted in P with probability at least $1 - \epsilon$ (i.e., the limit of the acceptance probabilities is 1).

**Reduction.** Given a PFA $\mathsf{P} = (S, \mathcal{A}, \delta, F, s_0)$, we construct a POMDP $G' = (S', \mathcal{A}', \delta', \mathcal{O}', \gamma', s'_0)$ with a reward function $\mathsf{r}'$, such that P satisfies the value 1 problem iff there exists an infinite-memory almost-sure winning strategy in $G'$ for the objective $\mathsf{LimAvg}_{=1}$. Intuitively, the construction adds two additional states good and bad. We add an edge from every state of the PFA under a new action \$, this edge leads to the state good when played in a final state, and to the state bad otherwise. In the states good and bad we add self-loops under a new action #. The action \$ in the states good or bad leads back to the initial state. An example of the construction is illustrated with Figure 3. All the states belong to a single observation, and we will use Boolean reward function on states. The reward for all states except the newly added state good is 0, and the reward for the state good is 1.

The key proof ideas for correctness are as follows:

*(Value 1 implies almost-sure $\mathsf{LimAvg}_{=1}$).* If P satisfies the value 1 problem, then there exists a sequence of finite words $(w_i)_{i \geq 1}$, such that each $w_i$ is accepted in P with probability at least $1 - \frac{1}{2^{i+1}}$. We construct an infinite word $w_1 \cdot \$ \cdot \#^{n_1} \cdot w_2 \cdot \$ \cdot \#^{n_2} \cdots$, where each $n_i \in \mathbb{N}$ is a natural number that satisfies the following condition: let $k_i = |w_{i+1} \cdot \$| + \sum_{j=1}^{i}(|w_j \cdot \$| + n_j)$ be the length of the word sequence before $\#^{n_{i+1}}$, then we must have $\frac{n_i}{k_i} \geq 1 - \frac{1}{i}$. The construction ensures that if the state bad appears only finitely often with probability 1, then $\mathsf{LimAvg}_{=1}$ is ensured with probability 1. The argument to show that bad is visited infinitely often with probability 0 is as follows. We first upper bound the probability $u_{k+1}$ to visit the state bad at least $k+1$ times, given $k$ visits to state bad. The probability $u_{k+1}$ is at most $\frac{1}{2^{k+1}}(1 + \frac{1}{2} + \frac{1}{4} + \cdots)$. The above bound for $u_{k+1}$ is obtained as follows: following the visit to bad for $k$ times, the words $w_j$, for $j \geq k$ are played; and hence the probability to reach bad decreases by $\frac{1}{2}$ every time the next word is played; and after $k$ visits the probability is always smaller than $\frac{1}{2^{k+1}}$. Hence the probability to visit bad at least $k+1$ times, given $k$ visits, is at most the sum above, which is $\frac{1}{2^k}$. Let $\mathcal{E}_k$ denote the event that bad is visited at least $k + 1$ times given $k$ visits to bad. Then we have $\sum_{k \geq 0} \mathbb{P}(\mathcal{E}_k) \leq \sum_{k \geq 1} \frac{1}{2^k} < \infty$. By Borel-Cantelli lemma [6, Theorem 6.1, page 47] we have that the probability that bad is visited infinitely often is 0.

*(Almost-sure $\mathsf{LimAvg}_{=1}$ implies value 1).* We prove the converse. Consider that the PFA P does not satisfy the value 1 problem, i.e., there exists a constant $c > 0$ such that for all $w \in \mathcal{A}^*$ we have that the probability that $w$ is accepted in P is at most $1 - c < 1$. We will show that there is no almost-sure winning strategy. Assume towards contradiction that there exists an infinite-memory almost-sure winning strategy $\sigma$ in the POMDP $G'$; and the infinite word corresponding to $\sigma$ must play infinitely many sequences of #'s to ensure $\mathsf{LimAvg}_{=1}$. Let $X_i$ be the random variable for the rewards for the $i$-th sequence of #'s. Then we have that $X_i = 1$ with probability at most $1 - c$ and 0 otherwise. The expected $\mathsf{LimAvg}$ payoff is then at most: $\mathbb{E}(\liminf_{n \to \infty} \frac{1}{n} \sum_{i=0}^{n} X_i)$. Since $X_i$'s are non-negative measurable function, by Fatou's lemma [6, Theorem 3.5, page 16]

$$\mathbb{E}(\liminf_{n \to \infty} \frac{1}{n} \sum_{i=0}^{n} X_i) \leq \liminf_{n \to \infty} \mathbb{E}(\frac{1}{n} \sum_{i=0}^{n} X_i) \leq 1 - c.$$

It follows that $\mathbb{E}^\sigma(\mathsf{LimAvg}) \leq 1 - c$. Note that if the strategy $\sigma$ was almost-sure winning for the objective $\mathsf{LimAvg}_{=1}$ (i.e., $\mathbb{P}^\sigma(\mathsf{LimAvg}_{=1}) = 1$), then the expectation of the $\mathsf{LimAvg}$ payoff would also be 1 (i.e., $\mathbb{E}^\sigma(\mathsf{LimAvg}) = 1$). Therefore we have reached a contradiction to the fact that the strategy $\sigma$ is almost-sure winning for the objective $\mathsf{LimAvg}_{=1}$.

**Theorem 5.** *The problem whether there exists an infinite-memory almost-sure winning strategy in a POMDP with the objective $\mathsf{LimAvg}_{=1}$ is undecidable.*